\documentclass[10pt]{article}
\usepackage[explicit]{titlesec}
\usepackage{hyphenat}
\usepackage{ragged2e}
\RaggedRight

\usepackage{rotating}  
\usepackage{theorem}   
\usepackage{bm}  
\usepackage{amsmath,amsfonts,amssymb}
\usepackage{booktabs}  
\usepackage{pdfsync}   
\usepackage{graphicx}
\usepackage{latexsym}
\usepackage{amsmath}
\usepackage{amssymb}
\usepackage{fancyvrb}
\usepackage{subfigure}
\usepackage{colortbl}
\usepackage{enumerate}
\usepackage{pifont}
\usepackage{stmaryrd}
\usepackage{textcomp}
\usepackage{fncylab}
\usepackage{multirow}
\usepackage{paralist}
\usepackage{wrapfig}
\usepackage{longtable}
\usepackage[]{algorithm2e}
\usepackage[table]{xcolor}
\usepackage{lscape}
\usepackage{lipsum}

\usepackage{garamondx}
\usepackage[T1]{fontenc}
\usepackage{amsmath}
\usepackage{graphicx}
\usepackage{xcolor}

\usepackage{geometry}
\usepackage{fancyhdr}
\usepackage[labelfont={footnotesize,bf} , textfont=footnotesize]{caption}
\makeatletter
\renewcommand\tagform@[1]{\maketag@@@ {\ignorespaces {\footnotesize{\textbf{Equation}}} #1.\unskip \@@italiccorr }}
\makeatother
\titleformat{\section}
  {\normalfont}{\thesection}{1em}{\MakeUppercase{\textbf{#1}}}
\titlespacing\section{0pt}{0pt}{-10pt}
\titleformat{\subsection}
  {\normalfont}{\thesubsection}{1em}{\textit{#1}}
\titlespacing\subsection{0pt}{0pt}{-8pt}

\makeatletter
\newcommand\sixteen{\@setfontsize\sixteen{17pt}{6}}
\renewcommand{\maketitle}{\bgroup\setlength{\parindent}{0pt}
\begin{flushleft}
\sixteen\bfseries \@title
\medskip
\end{flushleft}
\textit{\@author}
\egroup}
\makeatother

\usepackage[sort&compress]{natbib}
\makeatletter
\renewcommand\@biblabel[1]{\textbf{#1.}\hfill}
\makeatother

\bibpunct{}{}{,~}{s}{,}{,}
\title{Towards Artificial Intelligence Enabled Financial Crime Detection}

\author{
Zeinab Rouhollahi*$^{a}$ \\ \medskip
$^{a}$Macquarie University, Sydney, Australia \\  \medskip
zeinab.rouhollahi@students.mq.edu.au
}

\begin{document}

\vspace*{.01 in}
\maketitle
\vspace{.12 in}

\section*{abstract}

Recently, financial institutes have been dealing with an increase in financial crimes. In this context, financial services firms started to improve their vigilance and use new technologies and approaches to identify and predict financial fraud and crime possibilities. This task is challenging as institutions need to upgrade their data and analytics capabilities to enable new technologies such as Artificial Intelligence (AI) to predict and detect financial crimes.
In this paper, we put a step towards AI-enabled financial crime detection in general and money laundering detection in particular to address this challenge.
We study and analyse the recent works done in financial crime detection and present a novel model to detect money laundering cases with minimum human intervention needs.

\section*{keywords}
Financial Crime Detection; Feature Engineering; Process Data Analytics

\vspace{.12 in}


\section{Introduction}

Financial crime is an unlawful act and may involve various types such as fraud (e.g., cheque fraud and credit card fraud) and money laundering.
Recently, financial institutes have been dealing with an increase in financial crime. For example, as reported by Forbes\footnote{https://www.forbes.com/sites/steveculp/2020/08/26/why-banks-need-to-sharpen-their-focus-on-financial-crime/}, in April 2020 United Kingdom experienced a 33\% increase in Financial crime and fraud. Similarly, in the United States, IC3\footnote{https://www.ic3.gov/} (Internet Crime Complaint Center) reported as many fraud reports (by the end of May 2020) as it had in all of 2019.
This could result from the global shift towards a digital society due to COVID-19\footnote{https://en.wikipedia.org/wiki/COVID-19 pandemic} pandemic.

In this context, financial services firms should improve their vigilance and use new technologies and approaches to identify and predict financial fraud and crime possibilities. This task is challenging as institutions need to upgrade their data and analytics capabilities to enable new technologies such as Artificial Intelligence (AI) to predict and detect financial crimes~\cite{DBLP:conf/birthday/YangTB21}.
In this paper, we put a step towards AI-enabled financial crime detection in general and money laundering detection in particular to address this challenge.
We study and analyze the recent work done in financial crime detection and present a novel model to detect money laundering cases with minimum human intervention needs.

The rest of the section organized as follows:
In Section~\ref{overview}, we present an overview of this paper.
We present the problem statement in Section~\ref{problem}.
To better understand the research problem, in Section~\ref{motivating}, we present a motivating scenario in money laundering.
In Section~\ref{Contributions}, we present the contributions of this paper before concluding the Section with the summary and the paper outline in Section~\ref{Summary}.


\subsection{Background and Overview}
\label{overview}

In recent years, several regulatory authorities have increased their attention on detecting and preventing financial crimes. This is specifically important for banks and financial institutions as they have several business processes~\cite{DBLP:journals/spe/BeheshtiBM18,DBLP:books/sp/BeheshtiBSGMBGR16,DBLP:conf/caise/BeheshtiBN13} and procedures that are a good source for criminal actors.
Among these processes, we can mention cash depositing, overseas money transfer, and loan re-payment. If specific procedures occur in all banks and financial institutions internationally, then the amount of money available to commit financial crimes will decrease; therefore, putting specific procedures to recognize and prevent criminal activities is one of the challenges that banks face.


Several AI-enabled techniques aim to detect financial crimes in different sectors such as the bank and insurance industry, commodity exchange, security systems, stock markets, and money laundering to address these challenges. Emerging research and technologies which use big data analytics have helped significantly in collecting data and analyzing it. This, in turn, has led to a shift from traditional data collection methods to computer-based data gathering and analysis. This means that a vast amount of data being gathered and stored in each area, and organizations need proper strategies for analyzing these data efficiently and appropriately~\cite{Jantan2018State-of-The-ArtSurvey}.
Traditional approaches of detecting financial crimes would mainly lead police forces to look for so-called "street crime". In contrast, the development of new techniques using machine learning has made it possible to have a deeper look and search for high-level financial crimes committed by "white-collar criminals" as well~\cite{CliftonPredictingArsenal}.

\subsection{Key Research Issues}
\label{problem}

Financial Crimes have a widespread effect on economic, political, and social aspects of societies both in the national and international dimensions of society~\cite{Achim2020EconomicCrime}. While putting specific procedures for preventing, detecting, and reporting related cases has been an ever-growing concern of the banks, failing to do so correctly may lead to irrecoverable reputation losses for banks. Moreover, it may lead to a large amount of loss due to being fined by regulators and related authorities.
In particular, financial crimes may have a significant impact on the economy. For example, it may lead to challenges between political parties. Lack of proper regulations in financial markets may have two effects: (i)~it may avail new opportunities for financial crimes; and (ii)~it may endanger the health and steadiness of international financial structures~\cite{Masciandaro2004GlobalCentres}.

Considering new methods of financial exchange and the amount of financial data generated daily, it has become almost impossible to rely on traditional methods for detecting financial crimes. In this context, extracting relevant data from documents (such as PDF forms~\cite{DBLP:conf/adc/RastanPSRB18}) and analyzing crime patterns is considered to be a key challenge for law organisations~\cite{2004CrimeExamples}.
In this context, Artificial Intelligence (AI) and machine learning can enable machines to comprehend and learn to facilitate the detection and prediction of financial crimes.
In this paper, We study and analyze the related work in AI-enabled crime detection and present a novel model to detect money laundering cases with minimum need for human intervention.

 \subsection{Motivating Scenario}
\label{motivating}

Financial crimes can happen in different ways, such as money laundering, fraud, electronic crime, terrorist financing, and bribery and corruption.
Money laundering is known as the illegal process of concealing the source of money which has been obtained through criminal activities and putting this money into legitimate financial systems~\cite{MuhammaddunMohamed2012InvestigationMalaysia}.
In recent years, several works have been done in money laundering aiming to detect suspicious transactions. However, the need for a practical model which takes into account the current features of transactions that are happening in banks on a day-to-day basis and takes into account all aspects of each transaction is obvious.

A significant problem in this area is that the number of cases reported but not a money laundering case is high, which means that much time and effort is wasted in recognizing the unusual activities that are not a money laundering case.
%
In this paper, we present a novel model to detect money laundering cases with minimum need for human intervention to address this challenge. This model will use both classification and anomaly detection to detect cases of money laundering.

\subsection{Contributions Overview}
\label{Contributions}

Current models used by banks may have two problems. Firstly, they are mainly rule-based; therefore, not all of the cases covered, and the rules might miss some of the cases.
Secondly, current models have a high rate of false positives, meaning that the systems will alert for the cases that are not money laundering cases; therefore, too much time and effort could waste each day.
Our proposed model aims to cover these deficiencies
using a hybrid model where both classification and anomaly detection will apply to transaction data. The goal is to intelligently combine the two models to ensure that while most money laundry transactions are detected, there are not too many false-positive cases reported to decrease the amount of the effort done and increase the model's efficiency.

\subsection{Summary and paper Outline}
\label{Summary}

This Section discussed the background, key research issues, and contributions overview of this paper.
The remainder of this paper organized as follows.
\textbf{Section 2:} This Section will have a comprehensive look at the literature related to financial crimes and money laundering. We will investigate current methods used to detect financial crimes and specifically money laundering. We will have a deeper look at the machine learning methods used in this area and categorize different AI-enabled methods to detect money laundering.
\textbf{Section 3:} This Section will present the methodology that we are using in this research. It will describe the structure of the data and the features used for transactions. The classification model and anomaly detection that we apply to the data will also be described here.
\textbf{Section 4:} In this Section, we describe the outcome of the model applied to our dataset. The tools and techniques used for the implementation and the results, including the model's accuracy, are presented here.
\textbf{Section 5:} In this Section, we conclude the paper by reviewing the results and looking into future areas for improvement.

\section{Background and State-of-the-Art}

\subsection{Financial Crimes}
In today's competitive financial market, detecting fraudulent financial activities is one of the main success requirements of any financial organization. Banks and financial institutions are continuously working on novel and intelligent applications to discover such illegal activities because of the regulatory obligations and the extraordinary importance of obtaining a positive reputation amongst customers and achieving their trust.
Financial crime includes different types of fraud; examples are fraud via check and credit card, health care card, and point of sales. The act done in these crimes involves tax violations, identity thefts, cyber-attacks, and money laundering. In the next section, we will dive deeper into the definition and implications of money laundering.

\subsubsection{Money Laundering}

Money laundering is known as an illegal process that happens through banks and other financial channels. In this process, large amounts of money are transferred, and the source of the money is criminal but is concealed through several complex banking transactions and commercial money transfers~\cite{Gao2007AResearch}.
There can be several criminal activities associated with money laundering; this means that the criminals committing the crimes (such as smuggling, drug sales, forgery, terrorism, fraud, human trafficking, tax evasion, child laboring, and child sexual abuse) will conceal the sources of their money using banking and financial transactions to disguise the sources of their crime~\cite{Demetis2018FightingUK}.

\subsubsection{Increasing Money Laundering Risk}

The growing use of digital channels and the invention of digital money concepts such as bitcoin have added to the risk of the money sources concealed throughout these channels. A recent study shows that the use of Internet sources and the web has helped criminals launder the sources of criminal money~\cite{vanWegberg2018BitcoinBitcoin} which mean that in the upcoming years, with the invention of new money concepts and increased use of the Internet, the amount of money Laundering could potentially increase.

The amount of money that was transferred in money Laundering is not known precisely. However, it is estimated to be between 2 to 5 percent~\cite{Buchanan2004MoneyObstacle, Tiwari2020AAreas} of the overall Gross Domestic Product (GDP) throughout the world, which is a considerable amount compared to other financial activities. So the governments need to find and stop these activities. For instance, during the year 2016, 8776 money Laundering cases have been reported. Among these transactions, 72 percent reported by the banks and 23 percent by PSPs (Payment Solution Providers)~\cite{20186SUMMARY}. Considering the above figures, governments should monitor these activities and prevent criminals from committing these types of crimes by correctly detecting these activities.

Globalization and the fact that the financial markets in different countries have been brought closer alongside the ability to do online transactions 24 hours a day have increased the risk of Money Laundering. The ability to invest in other countries has made the opportunity to do what may not be possible in one's country and through clear financial systems~\cite{Buchanan2004MoneyObstacle}.

On the other hand, it is tough to recognize money laundering activities. Large amounts of daily transactions and different and complicated methods that money launderers will use to conceal the sources of data make it hard to investigate these activities. Vast amounts of data and multiple legacy systems make it hard for banks to recognize money Laundering activities appropriately. Further to that, several countries, especially among the developing countries, do not follow international rules and regulations on financial activities, making it a good place for criminals to hide the origins of their money throughout several transactions and banking activities. Furthermore, money launderers are changing their behaviors each day, making it almost impossible to place specific rules to prevent those activities because they become aware of the rules soon and find other mechanisms for money laundering~\cite{Bustan2008IntelligentSystem}.

\subsubsection{Controlling Money Laundering}

Several international authorities are in charge of controlling and preventing money Laundering. For example, FATF, the Financial Action Task Force (on Money Laundering), is an intergovernmental organization founded in 1989 and aims to develop policies to fight money laundering and financial terrorism activities. APG, the Australia Pacific Group on money laundering, is also an organization aiming to implement international standards to fight against money laundering, financial terrorism, and financing weapons.

In Australia, AUSTRAC, the Australian Transaction Reports and Analysis Centre, is the government agency responsible for identifying and preventing money laundering, tax evasion, fraud, and terrorism financing. In recent years, several rules and regulations were imposed on banks and financial institutions in Australia to know their customers and prevent money laundering activities.

The criminal actors involving in money laundering include criminal actors and professional money launderers (PMLs). These PMLs are the people who act in a professional capacity, such as a lawyer or accountant but knowingly get involved in the money laundering process to conceal the sources and show the criminals the methods to escape legal sources~\cite{20186SUMMARY}.

Money launderers would use several tools to disguise the sources of money. Several institutions and different methods of transferring money are available, so it is easy to wash the money. Further to that, as they become aware of the tools and patterns that banks are recognizing, they change their methods for concealing the source of their money~\cite{GaoIntelligent}.

In order to be able to prevent money laundering, we need to know how it takes place. Generally, the Money laundering process includes three steps: Placement, Layering, and Integration. Placement includes placing the money obtained from criminal activity into banks and financial systems by depositing the money through different geographical locations and legitimate sources. Layering contains concealing the sources of the money by doing many transactions and involving offshore accounts and complex investment vehicles. In the Integration phase, money launderers will transfer the funds to the owners or criminal actors. This is often done as a form of investment or tangible goods
like luxury cars and jewellery~\cite{Savage2016DetectionNetworks}.

\subsection{Money Laundering Detection}

Detecting money laundering activities and preventing them is an essential task of both governments and financial institutions. In recent years, several new regulations have been imposed on financial institutions to report any suspicious activity which could potentially be a source of money laundering. Banks and financial institutions have been obliged to put certain processes and procedures in place to report and detect any activity which can lead to money laundering~\cite{Singh2019Anti-MoneyActivity}.

In the academic area, several studies have been done, and this subject has been to the attention of scholarly published papers. In a recent study, six areas have been defined for the works that have been done in the field of money laundering, which include "anti-money laundering and its effectiveness, effects of money laundering on the economy and other fields, the role of different actors in this area and its importance, the magnitude of money laundering, and new opportunities for money laundering and its detection"~\cite{Tiwari2020AAreas}.

The solutions used by financial institutions are mainly based on statistical variables like mean and standard deviation. This is not efficient because it consumes a lot of effort from humans to detect suspicious cases~\cite{LeKhac2010ApplicationStudy}. Increasing precision is one of the main challenges in the area of money laundering detection. Each day, several transactions are being generated through different banking channels, and the rule-based software systems used in the bank recognize a significant number as unusual. In contrast, the detected unusual contain lots of false-positive cases that need further human investigation to distinguish true-positive from false-positive transactions. This final step is defined as recognizing "Suspicious" transactions out of "Unusual" ones. Therefore, increasing precision is an essential aspect that some researchers have looked into~\cite{LeKhac2010ApplicationStudy}.

It is worth adding that an "Unusual" transaction can be any transaction that does not follow the regular pattern of the rest of the transactions. This can be a high amount of money being transferred in one transaction or transferring money to a country under sanctions. However, just being "Unusual" does not mean that this is a "Suspicious" transaction that has been used for criminal activities. If a transaction is known as "Unusual" in bank systems, it will go into human investigations. Suppose it is recognized as "Suspicious". In that case, it will be sent to AUSTRAC along with other information on the account and account holder to decide whether it was a financial crime or not.
Based on a model presented by Wang et al.~\cite{Bustan2008IntelligentSystem}, the overall Anti-Money laundering (AML) process which is performed in banks and financial institutions can be described as below steps:

\textbf{Intelligence}: In the intelligence phase, we need to communicate with legacy systems to see the current rules and regulations proposed for detecting money laundering. We also need to have accurate information on accounts and their everyday transactions.

\textbf{Design}: In the design phase, we need to assess whether behavior in an account is usual or unusual. This can be decided based on the amount and type of the transaction or whether there was a movement in the historical average amount of the transactions.

\textbf{Choice}: In the choice phase, we need to make a warning report if any of the parameters in the design phase were not expected. We also need to review the transaction history to see what the behavior was during the time.

\textbf{Review}: In this stage, human investigators and experts from the bank will assess whether the suspicious behavior was a money Laundering activity or it was just an unusual behavior of the account.


\subsubsection{Anti-money Laundering: State of the Art}

\begin{figure}[t]
    \centering
    \includegraphics[width=0.9\textwidth, angle=0]{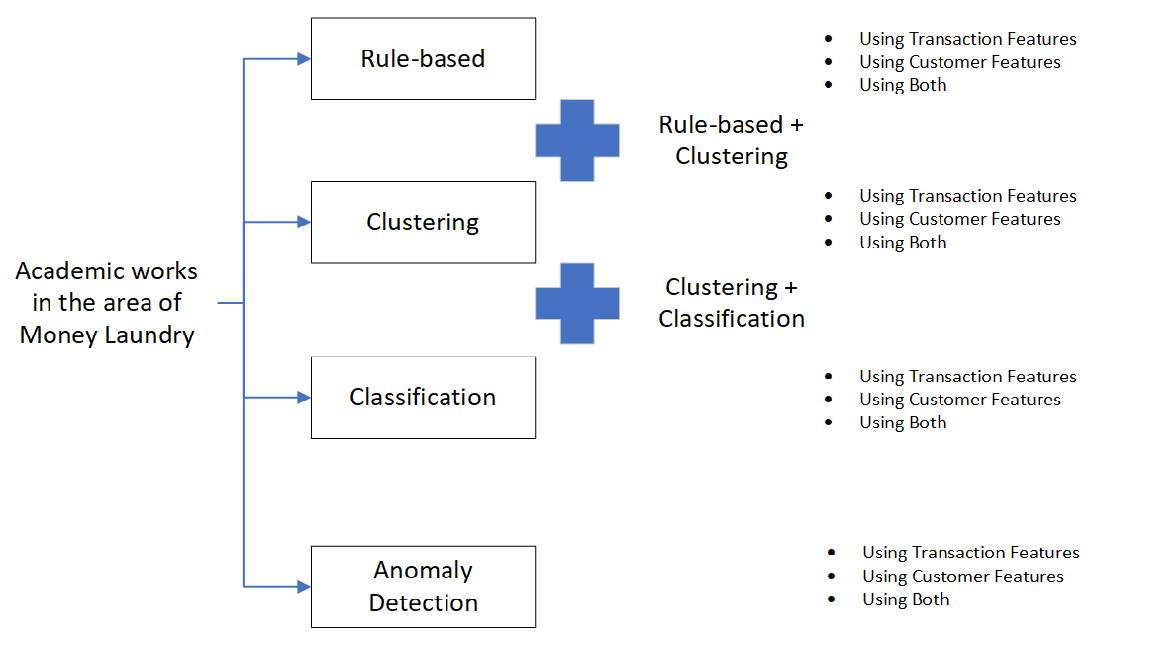}
    \caption{Categorizing Intelligent Methods in the area of Anti-Money Laundering.}
    \label{fig: Categorizing works done in AML.png}
\end{figure}

As it can be seen in Figure~\ref{fig: Categorizing works done in AML.png}, we can categorise different works, research methods, and machine learning applications in the domain of money Laundering detection into unsupervised approaches, including clustering methods such as~\cite{Wang2009ResearchApplication,CaoDangKhoaandDo2012ApplyingIndustry}, anomaly detection applications such as~\cite{Zhu2006AnSurveillance}, and rule-based models such as~\cite{Khan2013AReporting,Rajput2014OntologyDetection}, as well as supervised applications including classification and prediction methods such as ~\cite{Lv2008ALaundering,Tang2005DevelopingSVM,Salehi2017DataLaundering,DBLP:conf/edoc/JalayerKBPM20}.

In the rule-based models, as will be described in detail in the following subsection, we will identify unusual transactions for further investigation by assigning some rules on transaction features. Such rules can be numeric thresholds on features like transaction amounts and binary flags such as specific sources and destinations. The other approach is clustering, which segments customers and transactions based on their similarities in their features, making it more efficient to dive into each segment and analyze the transactions of each segment.
As shown in Figure~\ref{fig: Categorizing works done in AML.png}, some researchers combine clustering and rule-based approaches and apply specific rules on the transactions of each cluster that are more similar to each other. On the other hand, some supervised applications aim to detect suspicious transactions based on the historical data of other transactions.

In classification models, researchers try to use different classifiers to predict the transaction based on training the model with a large dataset of labeled transactions, i.e., transactions with some features labeled as suspicious or non-suspicious. There are also some hybrid works by a combination of clustering and classification models that aim to label transactions within each cluster. Moreover, anomaly detection applications compare transactions with the average pattern of other transactions to detect unusual ones.  In the following sections, we will review some works within each of the mentioned categories.

\subsubsection{Rule-based Models}

To detect money Laundering transactions, the initial step of all banks and financial institutions is the extraction of so-called "Unusual" transactions, which will lead to discovering suspicious transactions. The software tools which are used for AML (Anti-Money Laundering) in banks are usually rule-based. This means that these software programs are designed based on pre-defined rules and thresholds of average and standard deviation to detect Laundering transactions. In other words, the conventional approach to detecting unusual transactions is applying some rules about a specific transaction attribute or a set of them~\cite{DBLP:conf/intellisys/SchiliroBM20}. These attributes may contain transaction features such as transaction type, amount, statement, time, location, frequency, origin, and destination. For instance, any transactions with an amount above a specific threshold might be extracted as unusual. In a similar pattern, transactions with some pre-defined origin or destinations such as FATF blacklisted countries might be detected as unusual. Another example is about any transaction that includes specific words in its statement like terrorism activity-related words, i.e., names of specific years or locations. There are also rules made by combining some features such as any cash transaction above a specific threshold in Australia or any repeated transaction to a specific account (a combination of frequency and destination)~\cite{MuhammaddunMohamed2012InvestigationMalaysia}.

\begin{figure}[t]
    \centering
    \includegraphics[width=0.35\textwidth, angle=0]{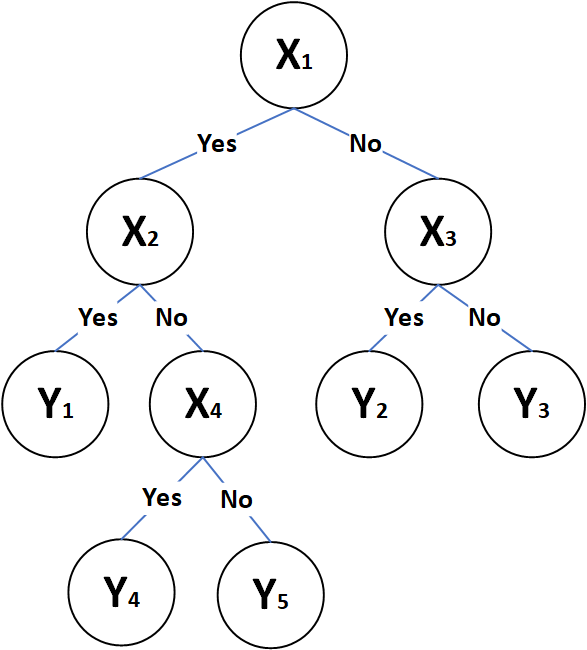}
    \caption{Decision tree rule-based money Laundering detection~\cite{Wang2007ATree}.}
    \label{fig: Decision Tree.png}
\end{figure}

Khan et al.~\cite{Khan2013AReporting} worked on detecting suspicious transactions by using statistical rules. They compared numeric transaction features such as amount and frequency with their average quantities. Whenever the deviation is above a certain amount, the rule will extract the transaction as a suspicious money laundering transaction. Rajput et al.~\cite{Rajput2014OntologyDetection} and Khanuja et al.~\cite{KhanujaHarmeetKaurandAdane2014ForensicTransactions} also set some rules on a combination of transaction features including amount, type, origin, and destination to detect suspicious transactions and then investigating these transactions by banking domain experts. One of the main tools and techniques to apply a set of rules to transactions is using the decision tree model shown in Figure~\ref{fig: Decision Tree.png}. For example, Liu et al.~\cite{Liu2011ResearchAlgorithm} applied this model on clusters that are provided by k-means to identify suspicious transactions within each cluster. Wang and Yang~\cite{Wang2007ATree} also applied a decision tree with some risk-related rules on the customer accounts.

\subsubsection{Clustering}

Several works used clustering models for detecting money laundering. Such models apply various clustering methods to the transactions and, after discovering the most efficient clusters of transactions, investigate each cluster as well as detecting the outliers~\cite{Salehi2017DataLaundering}. As there is a massive dataset of transactions for each customer and each transaction includes many features, we need a clustering method with a triumphant performance with large and multi-dimensional datasets~\cite{DBLP:conf/springsim/BeheshtiM07}. K-means is the most commonly used clustering method with excellent performance with such large and multi-dimensional datasets; however, the original version of k-means is not successful in detecting outliers. In other words, such outliers might deviate the resulted clusters, and therefore, there are some modified k-means methods designed to be able to detect the outliers~\cite{Ezenkwu2015ApplicationServices}.
Le Khac et al.~\cite{LeKhac2010ApplicationStudy} have combined natural computing and data mining techniques to propose a solution to detect some generic patterns for recognizing money laundering.

\subsubsection{Classification Models}

One of the main applications of machine learning in detecting fraudulent transactions and precisely transaction suspicious to money Laundering is classification. This application works by labeling different transactions with a binary label, identifying whether a transaction is suspicious of Laundering. Each transaction is converted to a vector model with multiple features. These features contain some numeric attributes such as transaction amount, frequency, and time. They also include other descriptive non-numeric attributes such as transaction type (cash/credit), source, destination, and method.

In some research, transaction features are used in the classification application, and customer features are considered. In other words, apart from transaction attributes, some other features contribute as the input of classification, which includes the customer credit score, customer average spending on specific domains. This includes gambling as well as socio-demographic features of the customer, including their age, income, education, etc~\cite{Heidarinia2014AnSystems}.
There are money laundering detection applications that use "Support Vector Machine", "Correlation Analysis", and "Histogram Analysis" to detect money laundering and identify money laundering cases. The dimensions used in these methods include "Customers", "Accounts", "Products", "geography" and "Time"~\cite{AnABank,Zhang2003ApplyingCrimes}.
Labeling can be hard for Laundering as the proportion of the laundering transactions is deficient compared to the whole transactions. A simple method for labeling a transaction as suspicious in supervised methods is selecting the suspicious transactions from those marked by domain experts. In contrast, general transactions which exist in the pool of transactions can just be assumed as usual because the portion of the suspicious transactions from all transactions is low~\cite{Savage2016DetectionNetworks}.

The data used in different studies vary extensively. Most researches in AML either use real small datasets or simulated data sets, and only a few researchers have applied the model on large real data sets. Some researchers study a specific type of transaction as money Laundering cases, for example, international fund transfers or a large amount of cash deposits~\cite{Jullum2020DetectingLearning}.
In a recent study by IBM corporation and MIT university, classification was used to detect money Laundering. The model proposes that analyzing AML data is a classification problem. We need to classify a small number of illicit transactions in large massive datasets of all transactions. Each node transaction has 166 features, and 94 of them are presenting some local information about the transaction like time step, the number of inputs and outputs, transaction fees, output volumes, and the rest of the features comes from aggregating transaction features. This model uses a time series of bitcoin transactions, directed payment flows, and node features. The model applies a binary classification to illicit transactions and applies Logistic Regression, Random Forest, Multi-layer Perceptrons, and Graph Convolutional Networks on the data. The results show that Random Forest works best~\cite{Weber2019Anti-MoneyForensics,DBLP:conf/ijcnn/KhatamiNB0NZ20}.

\subsubsection{Anomaly Detection}


One of the approaches in detecting transactions suspicious to be money laundering is using the ML application of Anomaly Detection to recognize the transactions which are not expected and show deviation from the expected behavior of the transactions. In other words, the technique used for anomaly detection should be able to identify the unusual behavior of each customer account~\cite{Chen2018MachineReview}.
Anomaly detection is usually adequate to detect abnormal behavior or pattern~\cite{LiuIsolation-basedDetection}. In the case of financial frauds, this is even more useful because complicated schemes are used to avoid security protocols in detecting illegal transactions~\cite{Baltoiu2019Community-LevelLaundering}. This application works similar to an advanced rule-based model that aims to compare the features of entities with some statistical parameters of features, i.e., the average and standard deviation, and then extract the observation with more than a specific deviation as unusual. For instance, for transactions of a specific customer, the first average and variance of various attributes such as the amount and the number of transactions per day are being calculated. Then a threshold will be set, such as a 300 percent deviation. Then any transaction with a deviation more than the assigned threshold to the average or variance will be extracted as unusual. The mentioned attributes might be any numerical features such as amount, time, and frequency. Similarly, by setting higher thresholds, the model will extract suspicious transactions~\cite{Rohit2015ReviewFramework}.

In a new anomaly detection method called isolation forest or iforest, anomalies are detected through isolating instances. This method achieves this aim by working on the attribute values that are different from the regular instances and without calculating the distance and density measure~\cite{LiuIsolation-basedDetection}.

\subsection{Features Used in Detecting Suspicious Transactions}

In order to detect money Laundering, some academic papers have studied transaction features while others have studied both transaction and customer attributes. Using customer features will add value to the model, but it will usually incur privacy issues, making it hard to gather and use the data.
Jullum et al.~\cite{Jullum2020DetectingLearning} introduces a machine learning method for detecting suspicious money Laundering cases, and both transaction features and customer features are used in this work. In contrast, some other works have just worked on the transaction features. The critical point on selecting the transaction features versus customer features is the privacy issue. Customer features usually include data on the customer profile, which needs to be masked to prevent privacy policy issues. In contrast, this issue is not as critical in working with the transaction features.

\subsubsection{Transaction Features}

There are several features associated with each transaction type. Different works in this area have considered different features. While most of the work done on anti-money Laundering use transaction features~\cite{Lokanan2019DataTransactions}, some of them also use the data available on the customer~\cite{Bustan2008IntelligentSystem,Jullum2020DetectingLearning}. For example, Jullum et al.~\cite{Jullum2020DetectingLearning} propose a machine learning method for detecting suspicious money Laundering cases using both transaction and customer features. The model used in this paper trained to forecast the probability of a money Laundering case using the background data related to the sender and receiver of the transaction, their previous financial behavior, and the transaction history related to each of them.


\begin{figure}[t]
    \centering
    \includegraphics[width=0.7\textwidth, angle=0]{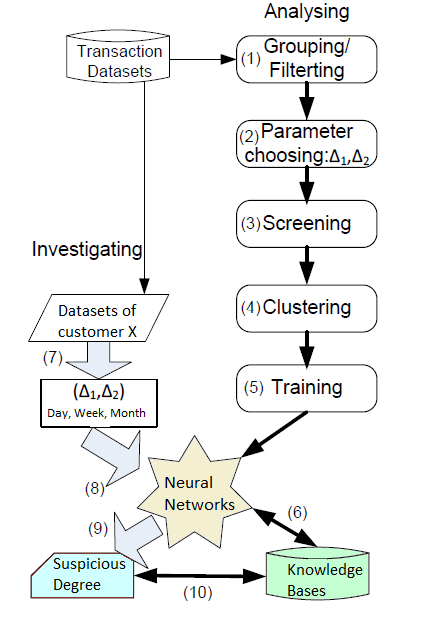}
    \caption{AML analyzing model~\cite{AnABank}.}
    \label{fig: Data Mining Framework}
\end{figure}

An-Le Khac et al.~\cite{AnABank} presented a model that tries to analyze transactions in the banks related to investment. Two essential characteristics in this analysis are the frequency of transactions and the value of each transaction. The model uses two parameters. One is the proportion between the redemption value and the subscription value conditional on time. The other one is the proportion between a specific value and the total value of an investor's shares conditional on time. The model then applies a clustering technique for these parameters on the fund level and investor level. The model will then use neural networks (back-propagation) and train them on suspicious and not suspicious cases. A decision tree is also built to train the model based on time. The results are then evaluated, and it is aligned with the needs of the AML unit. The overview of the model can be seen in Figure~\ref{fig: Data Mining Framework}.
The paper uses these steps to analyze the data:
(i)~\textbf{Data Pre-processing}: Extracting and cleaning-up the raw data sets and building a data warehouse of customers and transactions;
(ii)~\textbf{Data Mining}: Using clustering and classification techniques to analyze transactions: extracting the transaction records for investigation, associating multiple transactions to a specific account to know more about the financial activity of that account, building suspicious clusters of customers by clustering similar transactions, and classifying customers in pre-defined categories of risk; and
(iii)~\textbf{Knowledge Management}: Putting together the results of the mining process as well as the knowledge from the domain experts, results will be collected, stored, and analyzed~\cite{abu2021relational}. Also, interpretable rules and knowledge are generated at this stage.

There are several types of risks associated with a transaction in terms of money laundering, which can be investigated. Static risks are those related to the transaction and the account holder, while dynamic risk refers to the account's behavior during the time. Static risk includes risks related to the account holder (the person or company holding the account), type of the transaction or product, which can be cash or digital transfer, geography risk, which refers to the location source and destination of the transaction (e.g., FATF listed countries), or amount risk which refers to the amount of the transaction. On the other hand, dynamic risk refers to the unusual amount of fund movement related to that account, which refers to the short-term and long-term moving average of the account associated with the transfer~\cite{Bustan2008IntelligentSystem}.

\begin{figure}[t!]
    \centering
    \includegraphics[width=0.9\textwidth, angle=0]{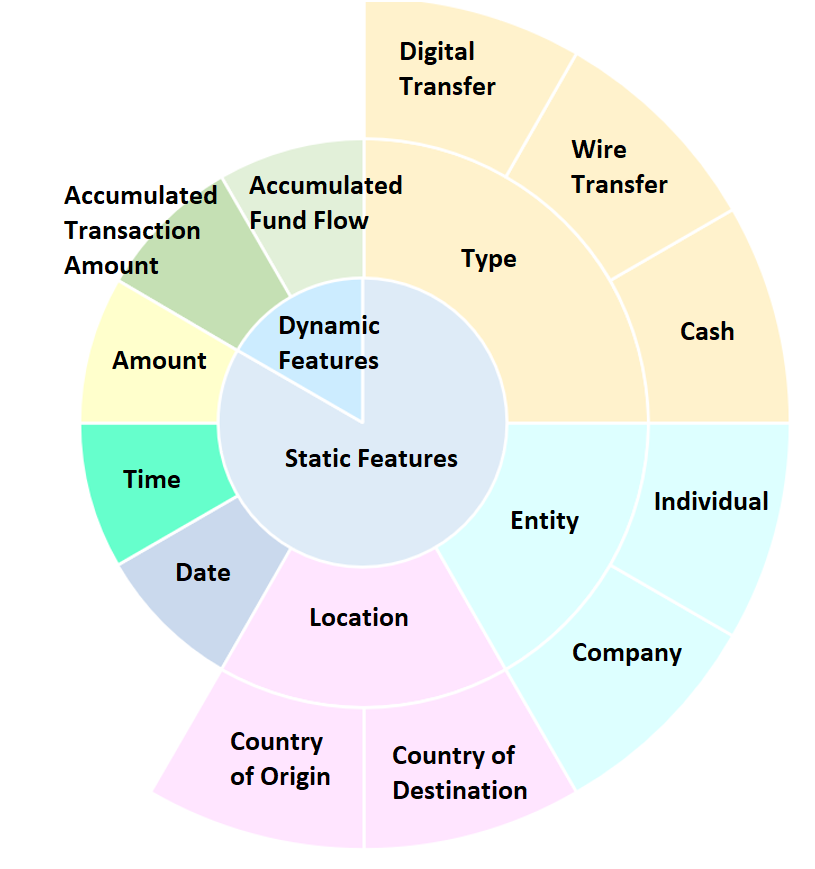}
    \caption{Transaction Features.}
    \label{fig:Transaction Features}
\end{figure}

In order to reduce these risks, we will associate each risk with its relevant transaction features and identify features of the transaction that will help us reduce these risks. Based on the risk model as well as the transaction features mentioned in the literature, features such as the time of the transaction, type of transaction (cash, money transfer, wiring), entity (Person or company doing the transfer, average person, or PEP), destination account, geographical location of the origin and the destination of the transaction, amount of transaction, accumulated fund flow, and accumulated transaction amount recognized as critical for money Laundering detection in the area of anti-money laundering and financial crime detection. The transaction features can be seen in Figure~\ref{fig:Transaction Features}.

\subsubsection{Customer Features}

While some works have just concentrated on transaction features, others have used some features associated with the customer or the account. These works have not just concentrated on the customer, but instead looked at both customer and transactions simultaneously and applied the models on both.
Le Khak et al.~\cite{AnABank} have used redemption value and subscription value and identified two new factors using the proportion between these features. In this research, there is no direct mention of the customer features, but each transaction's details are associated with the account detail related to each customer.
Sudjianto has mentioned transaction features such as the total amount of transactions, transaction type, frequency, average amount, and maximum during a time (for example, a week), and customer features such as customer profile, income amount, etc~\cite{Sudjianto2010StatisticalCrimes}.

\subsection{Labelling Methods}

Labeling the training data is one of the most important parts of machine learning models, i.e., classification applications used in money Laundering detection to increase different accuracy indexes such as accuracy, precision, recall, and F1-score. This labeled training data will play a significant role in the accuracy of the model. In other words, having not sufficient data, the model might lead to a low prediction accuracy which means the whole model will not be reliable enough.

Labeling training data is indeed one of the most significant obstacles to deploying machine learning systems. The simplest way to approach the labeling would be hand-labeling the training data; however, hand-labeling is expensive and time-consuming. It would help if we could find relevant people and, in the case of financial crimes, subject matter experts (SMEs) to label the data, which will be hard to find the resource and also time-consuming and expensive~\cite{Ratner2020Snorkel:Supervision}.
Creating the training data set is expensive, especially when it needs expert knowledge. Regarding an accurate money Laundering classification model, for example, we need a dataset of more than 100000 rows, i.e., the transactions with several features for each of them. A label identifies whether or not the transactions are suspicious to be a money laundering transaction. Labeling hundreds of thousands of transactions with so many features by the banking domain experts will be much time-consuming, expensive, and generally inefficient. Therefore, working on alternative methods for labeling these transactions by deriving insights from labels generated by an expert has a significant impact on the model efficiency.

There are several ways that companies use to label the data; some of them hire large groups of people to do this for them manually; or use classic techniques such as active learning~\cite{Settles2009ComputerSurvey}, transfer learning~\cite{Pan2010ALearning}, and semi-supervised learning. Other sources of labeling the data can include using crowd-sourcing methods for labeling the data~\cite{AssociationforComputingMachinery.29thSystems.} and setting rules and heuristics for labeling ~\cite{Rekatsinas2017HoloClean:Inference,Ratner2020Snorkel:Supervision}.
The challenges of hand labeling the data are not limited to being expensive and time-consuming; other challenges also exist. Different sources of labeling the data can be conflicting and have different results. Also, the accuracy of some labeling methods is unclear, which makes it hard to estimate the overall accuracy of the label~\cite{Ratner2020Snorkel:Supervision}. Methods such as crowd-sourcing have their problems as tasks have non-boolean responses and workers have biases on positive and negative tasks~\cite{Joglekar2015ComprehensiveAlgorithms}.

\begin{figure}[t]
    \centering
    \includegraphics[width=0.9\textwidth, angle=0]{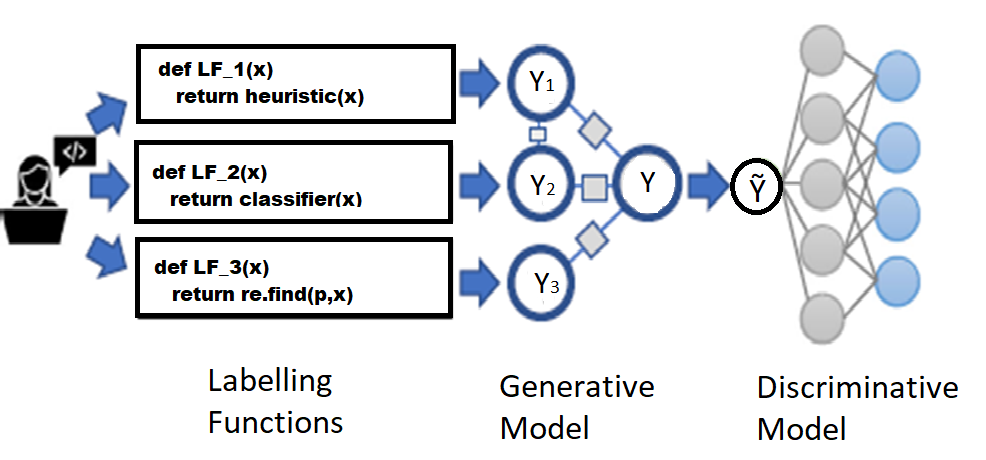}
    \caption{Snorkel Labelling Model Overview~\cite{Ratner2020Snorkel:Supervision}.}
    \label{fig: Snorkel Model}
\end{figure}

Some labeling methods generate probabilistic training labels using the agreements and disagreements between various sources of labeling the data. Specifically, in a model called Snorkel, developed at Stanford University, the labels will be determined based on the result of several simple labeling functions written by subject matter experts, and the model will determine the weight of each labeling function based on the agreements and disagreements between the labels for each data point~\cite{Ratner2020Snorkel:Supervision}.
An overview of the snorkel model can be seen in Figure~\ref{fig: Snorkel Model}.
The accuracy of this model is highly dependent on the accuracy of its labeling functions which consist of various rules, functions, or a set of rules and functions that automatically label the data based on their features. For example, in suspicious labeling transactions, based on the features extracted for each transaction, each labeling function might be a rule or a set of rules by assigning some thresholds to the features such as transaction amount, transaction statement,  origin, destination, customer type, credit, and cash. This model has an excellent performance in dealing with large datasets by applying accurate labeling functions~\cite{Ratner2020Snorkel:Supervision}.


\section{Methodology}

In this section, we will go through the methodology used in this research to detect and prevent money laundering cases.

\subsection{Method Overview}

\begin{figure}[t]
    \centering
    \includegraphics[width=0.8\textwidth, angle=0]{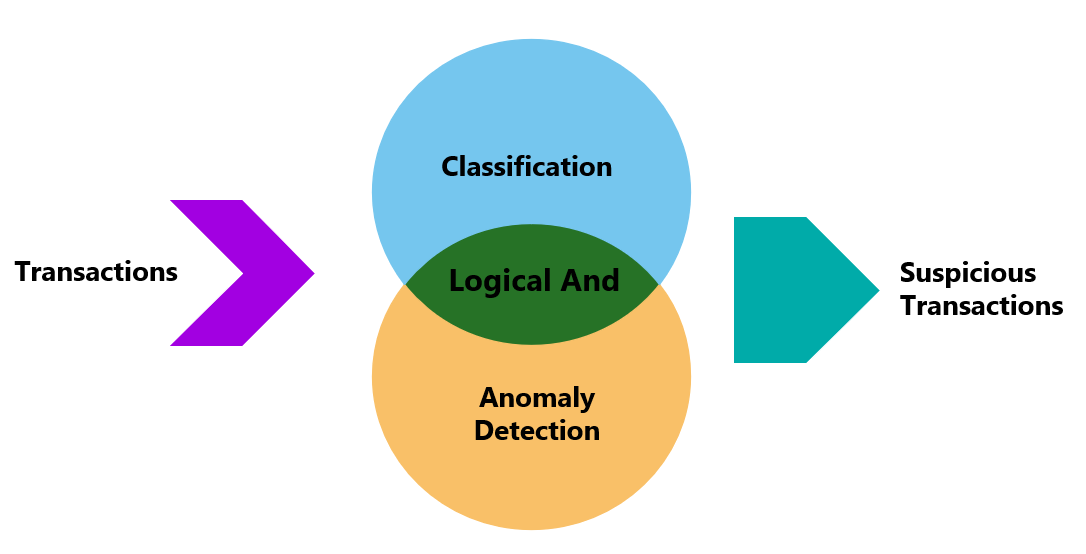}
    \caption{Applying Supervised and Unsupervised Methods to Detect Suspicious Transactions.}
    \label{fig: Supervised and Unsupervised.png}
\end{figure}

We leverage a novel intelligent hybrid pipeline including both supervised and unsupervised machine learning applications. We aim to decrease false-positive detection of suspicious transactions (i.e., the number of transactions that the model captures as suspicious but not a money laundering case will decline).
%
As illustrated in Figure~\ref{fig: Supervised and Unsupervised.png}, we aim to predict the suspicious transactions by training a supervised classification model in the first step. We also detect unusual transactions by training an unsupervised anomaly detection in the second step. Finally, by implementing a logical AND between the results, the model will detect unusual transactions as whatever is captured by both applications. This will lead to fewer suspicious detected transactions. As mentioned in section~2, one of the drawbacks of the state-of-the-art works of money laundering detection is the low precision of the results; we aim to focus on precision by applying the mentioned logical AND between the results of two applications.

We start with curating the dataset in a banking sector, which contains a set of transactions with specific attributes for each transaction.
We aim to label our transactions to demonstrate whether they are related to a money laundering case or not. In order to label the transactions, as will be explained in detail in Section~3.3.1, we are using banking analysts to help label part of the dataset. Then, we use learning techniques to label the rest of our large dataset automatically. We will discuss the labeling functions and auto-labeling methods in this section. Once we labeled the transactions, We will apply a classification model to detect suspicious transactions based on the labeled data and measure the accuracy results.  We will also apply Anomaly Detection to capture transactions' features that are unusual transactions and compare the results with the transaction labels to calculate the accuracy of anomaly detection applications, using the same indexes. Finally, we apply a logical approach between two classification and anomaly detection applications and measure the impact on the accuracy results.
The final goal is extracting suspicious transactions with minimum human intervention. Once we detect suspicious transactions, the bank will check other transactions of the same source and detect more transactions to be labeled by the analyst.

\subsection{Data Labelling}

Labeling the data is an essential step in our proposed approach. The accuracy of the model, including both classifications and anomaly detection applications, relies on the training data that we feed into it; therefore, wrong or misleading labeled data in the training data set might result in incorrect results. Having relevant labeled data is not always a straightforward task, as the actual sources for labeling the data might be inconvenient, and the replacing methods might not always be suitable. Therefore, finding and applying a proper method for labeling the data is critical in data-driven approaches. Especially in the banking industry, this will be more significant as the data is critical and not publicly accessible, making it even harder to access the properly labeled data.

We aim to use three types of data labels for money laundering cases. The first one is for detecting unusual transactions from the usual ones, which includes any transaction that banking systems will recognize as different from regular transactions. The second label is for detecting suspicious transactions from the non-suspicious ones. Bank financial crime experts currently do this step. It is a time-consuming task, and the current rule-based software solutions may not accurately identify relevant transactions.

Along with their account holder details, these suspicious transactions will then go to formal authorities, such as AUSTRAC, to identify money laundry cases. In this research, our labels are applied to the second type, referring to the "Suspicious" transactions. The first source of labeling data is banking experts. They identify suspicious transactions based on their expertise and experience. This type of labeling is timely and requires human effort, so the cost is high. According to the size of the dataset, we cannot label a dataset of 100k transactions by human expert labeling. Therefore, we use the proposed approach in addition to auto-labeling methods to identify suspicious transactions.

\subsubsection{Snorkel Model}

\begin{figure}[t]
    \centering
    \includegraphics[width=1.1 \textwidth, angle=0]{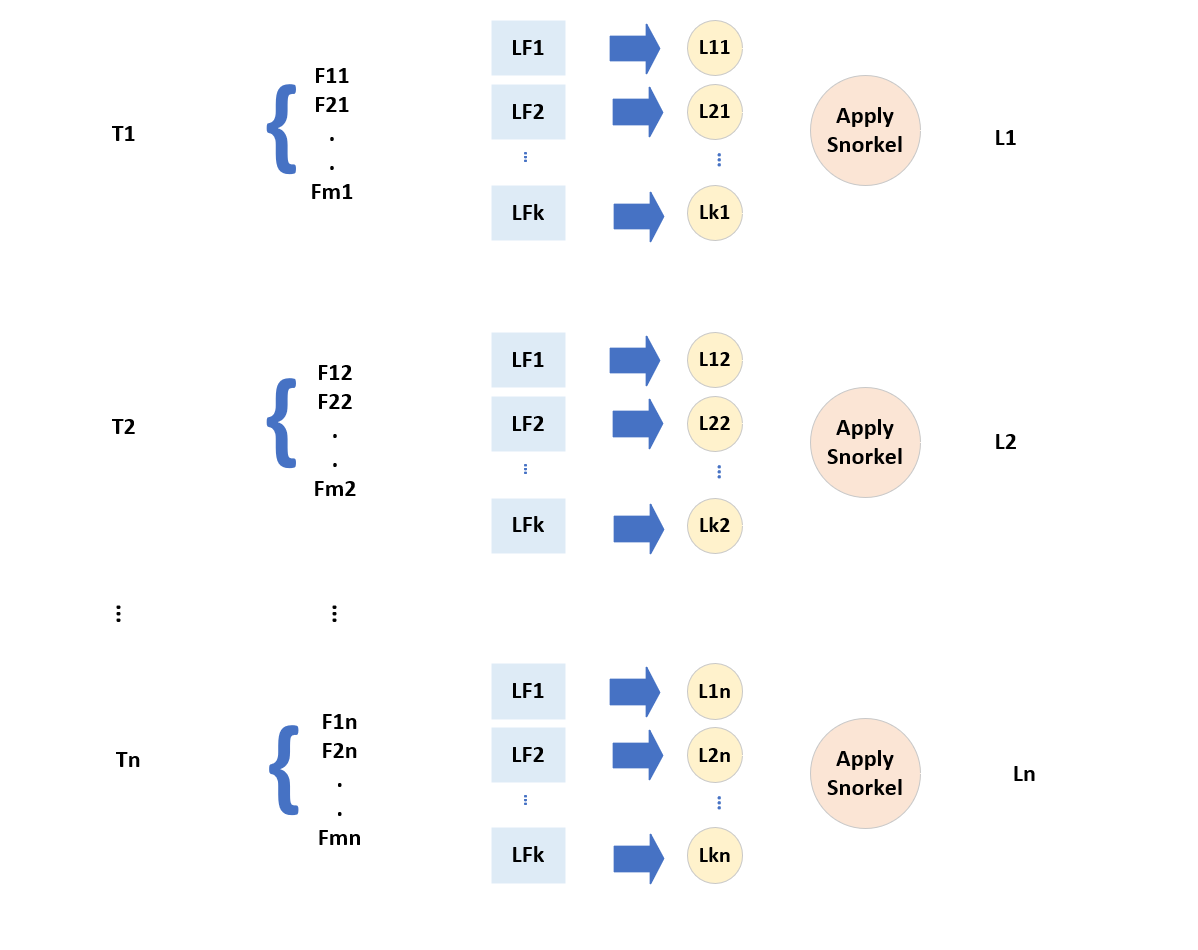}
    \caption{Applying Auto Labelling Techniques to Label the Training Data.}
    \label{fig: Auto Labelling the data.png}
\end{figure}

We are using a similar approach to the Snorkel model~\cite{Ratner2020Snorkel:Supervision} to facilitate auto-labeling the transactions.
The dataset used in this project contains 100,000 records consisting of transactions with features such as amount, source location, destination location, time of the transaction, and bank branch. To label this dataset, we are using banking experts for 10 percent of the dataset. For all the rest, we are applying the Snorkel model to label the data by using the simple rules as labeling functions. The Snorkel model is an auto-labeling method that creates the labels based on comparing the outcomes of different labeling functions, i.e., rules and selecting the labels according to the agreement of the majority of outcomes. Figure~\ref {fig: Auto Labelling the data.png} shows an overview of the model we have used in this project for labeling part of the training data.


To further explain this section, suppose that we have n transactions in a specific time duration called t1, t2, t3, ..., tn. Each transaction has a vector of different features like amount, time, country of origin, country of destination, product type, and transaction branch. These features can be seen below:

\emph{
V(t) = \{Transaction ID, Account ID, Product Type, Transaction Branch, Transaction Date/Time, Transaction Amount, Transaction Currency, Credit/Debit Status, Country of Origin, Country of Destination, Transaction source type\}
Each of these features has its definition, which is described in the previous section.
}

Suppose that each transaction has m features, named F1, .... , Fm.
This way, our Feature Vector for one transaction will be as followed:

$V(T1) = \{F11, F21, ...., Fm1\}$

$V(T2) = \{F12, F22, ...., Fm2\}$

...

$V(Tn) = \{F1n, F2n, ...., Fmn\}$

To auto-label the data, we will apply labeling functions to each of these transactions. We have k labeling functions, so LF1, LF2, .... , LFk are our labeling functions. We apply these labeling functions to each transaction. Snorkel model will find the labels of each transaction based on the agreement and disagreement between the labeling functions and will finally decide on a single label for each transaction. Each transaction will have labels as 1 for being a suspicious transaction and 0 for being a typical transaction. Therefore as demonstrated in Figure~\ref{fig: Auto Labelling the data.png}, applying auto-labeling on this data means finding the labels $L1, L2, ... , Lm$ for our transaction set. These labels are then used for training our model along with the labels from a banking domain expert. These labels are used as a supervised method for our classifiers. For anomaly detection, which is an Unsupervised method, we do not need the labels, as the model will recognize the anomalies without needing the labels. However, to compare the result of our anomaly detection model with the actual labels, we will then consider all anomaly transactions as suspicious ones and measure the accuracy, recall, precision, and recall of the model based on that.

\subsubsection{Labelling Functions}

In order to label the data based on the Snorkel model, we need some labeling functions. These labeling functions are written based on simple rules that can specify unusual transactions. For example, having a cash transaction with an amount more than a specific value will have a flag as an unusual transaction. The actual amounts will not be revealed in this paper explicitly to comply with the privacy policies of the sponsor company and regulations related to the financial crimes.

Regarding the unusual transaction, one crucial point is that if a transaction is identified as unusual, it does not mean that the transaction is money laundering. However, it is instead a way of flagging it to go for a different level of investigation. If other labeling functions agree with that, this transaction will be flagged as suspicious for money laundering.

Another labeling function is related to recognizing the money Laundering cases related to smuggling wildlife and animal skin. In this case, the country of origin or destination will be necessary. So if one of these countries is on a blacklist (which we cannot reveal their names because of privacy and are those who happen to be the primary source of wildlife trafficking), then the transaction will be flagged as unusual and needs to be passed for further investigation.

Another case of labeling functions would be for the money laundering cases related to terrorism activities. In such scenarios, if certain words are used in the transaction statement (e.g., a keyword related to the history of a terrorism activity or is a digit reminding the year of that terrorism event), then this transaction will be flagged as unusual and will be sent for more clarification via anti-money laundering teams. Examples of such words include "hijack" and "terror".

There are also labeling functions related to customer credit score. A credit score is a number between 0 to 1 (with 1 the most and 0 the least) and represents customers' credit. This number can be impacted by customer behavior such as loan activity, income, the total number of years that the customer has an account in the bank, the average amount of monthly transactions, expected lifetime value, and types of products used. In this study, from experience obtained from the domain knowledge experts, it has been recognized that money laundering activity usually happens within customers with a credit score less than 0.05. Accordingly, as a labeling function, a simple rule could be used to label a transaction as suspicious: if the customer credit score is less than 0.05 and the transaction amount is more than 20000. These rules are different from setting rules for detecting suspicious transactions, and they are just labeling functions that we will feed into the snorkel model. The model will decide if this is aligned with the labels identified by the expert and also the result from other labeling functions.

\subsection{Classification}

\begin{figure}[t!]
    \centering
    \includegraphics[width=0.8\textwidth, angle=0]{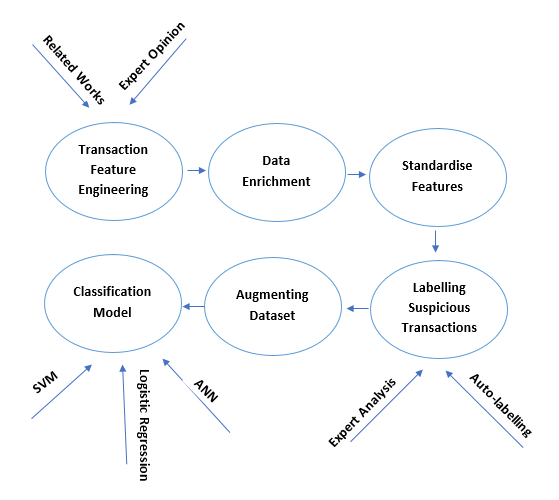}
    \caption{An Overview of the Classification Model for AML.}
    \label{fig: Model Overview.png}
\end{figure}

In this research, we apply the classification on transactions to predict whether they are suspicious and need to be considered money-laundering. Figure~\ref{fig: Model Overview.png} shows an overview of the classification method used in this research. The proposed model is a binary classification and aims to predict the suspicious transactions. Therefore we will have 1 as the label for suspicious transactions and 0 as the label for non-suspicious ones. The input of the classification model is the transaction features.

We can extract different features for a transaction, including account, time, type, entity, destination, location, amount, and fund flow statement. We are using the big banking data generated on open and private data islands. Using Knowledge-Bases, we enrich the raw data extracted for each customer to use them as input of our classifier. For instance, we are using Wikidata to enrich the words of transaction statements and derive some insights. For example, transactions with specific words (such as "Gambling" and Game") in their statement can be flagged.

Once we extracted the semantic features from our raw data about transactions, we feed them into a classification model. We need to consider some pre-processing on the data, including standardization, labeling, and Augmenting~\cite{DBLP:conf/birthday/BarukhZBBBCYSS21}. We will describe them below.

\textbf{Transaction feature engineering}: We first need to select important features which will help us identify money laundering transactions from regular transactions. These features are being selected through a pool of transaction attributes~\cite{DBLP:conf/momm/KhadivizandBSSI20,DBLP:conf/momm/RezvaniBT20}.

\textbf{Data enrichment}: One of the key challenges in data analysis is changing the raw data into curated data. To drive insight from the extracted data from all different sources with several formats, we need to use KBs to provide semantic data. For instance, to extract information from transaction statement we need to extract key-words and enrich the text by KBs that detect specific word groups~\cite{Beheshti2019DataSynapse:Foundry,Beheshti2020TowardsSystems}.

\textbf{Standardization}: As we have different numeric features with various ranges and different means, we need to do the mean normalization first to ensure the model will converge to the global minimum of its cost function.

\textbf{Labelling}:
While we are analyzing a large size dataset with thousands of transactions, suspicious transactions need to be detected by the analyst. Thus, we are not able to label all of the transactions used in training by the analyst. Instead, we need to label a portion of them by using intelligent techniques to automatically label the rest of the transactions by the patterns extracted from analyst labeling.

\textbf{Augmenting}: Our dataset is skewed, i.e., the portion of suspicious transactions is not equal to normal ones. Based on historical banking data, we expect a one-digit percentage for our suspicious transactions. Therefore, the labels are not balanced. In other words, we will have for example 10 percent of the transactions as suspicious, which are labeled 1 versus 90 percent labeled 0. However, to get the most accurate result from our classification, we need balanced labeling. That is why we are using augmentation techniques to balance our dataset according to the portion of suspicious transactions.

\textbf{Classification}: We use the pre-processed data as the input of classification application. We are using different classification methods and measure the results. In this research, we aim to use logistic regression, Nearest Neighbours, Naive Bayes, Neural Network, and Random Forest as our classifiers to predict the transactions that are categorized as money-Laundering. The evaluations of these models are being done by measuring Accuracy, Precision, Recall, and F1 Score for the classification result. As our dataset is skewed and has fewer positive labels (suspicious transactions) than normal ones, we need to use both accuracy and F1 score to evaluate our classification model.

\subsubsection{Transaction Features Vector}

The transaction features we have come up with so far in this paper contain the main attributes a transaction can have. These features will make a vector for each transaction and will consist of the below elements:

\emph{
Transaction ID, Account ID, Product Type, Transaction Branch, Transaction Date/Time, Transaction Amount, Transaction Currency, Credit/Debit Status, Country of Origin, Country of Destination, Transaction source type.
}

\subsection{Anomaly Detection}

The second phase that we are focusing on in this research is Anomaly Detection. For each transaction, we extracted different features as mentioned above. Then we will calculate the average and standard deviation for each to find the Gaussian distribution of each of the features. We will use our training data for statistical calculations, and we put regular transactions (not suspicious ones) as our training data. Then we will use a combination of regular and suspicious transactions as our cross-validation and test data to calculate the probability of each transaction. We will pick any transaction with less than a certain probability as suspicious. We will find the best threshold by checking different ones, finding the best F1 score within cross-validation data, and selecting the best threshold for the probability of unusual (suspicious) transactions.


We will use anomaly detection to extract more complex features. If a transaction is labeled as suspicious by the analyst, but the system does not identify it as suspicious, we will perform anomaly detection and identify more features; this can combine existing features and rerun the model with updated features. In each step, we will apply the anomaly detection method on cross-validation data and measure the accuracy and F1 score to find the most successful feature list and threshold.

\subsubsection{Applying Logical AND}

One of the main goals of this research is to minimize the need for human intervention to detect suspicious transactions. In other words, we need to minimize the number of transactions that are identified as suspicious while they are normal. So we need to decrease the portion of false positives and increase precision. One of the contributions of this research is capturing the agreement between the results of classification and anomaly detection to achieve this goal. Therefore, we will apply the two applications described above, and we will apply a logical "And" between them. The first application as a supervised machine learning model is a classification in which we need almost 100k labeled data. Hence by applying the Snorkel model and using expert knowledge, we can label the data and use it to train the model. The second application is an unsupervised one which is anomaly detection. We apply a logical AND between the results and measure the accuracy indexes to prove how this method will improve the precision according to the project goal.


\section{Experiment, Results, and Evaluation}


This section reviews the results of applying our methodology to real transactions to extract unusual and suspicious transactions. We first present our motivating scenario, which is about the detection of transactions suspicious of money laundering. We will then discuss our dataset specifications: the transactions of one of the Australian banks with the masked data according to the privacy policy. Each observation in this dataset is a transaction with its related features and attributes. Finally,  We will separately discuss the accuracy of each part of our proposed pipeline. We will provide the accuracy results of the classification model with the global indexes of accuracy for skewed datasets. Also, we will present the anomaly detection application results. Lastly, we discuss the impact of applying logical AND as part of our contribution on the accuracy results and will see how this method will assist in achieving the goal of this research. In our coding, we used Python 3.8, Pandas 1.2.4, and Numpy 1.20.2 to evaluate the results of our model.

\subsection{Motivating Scenario: AML in Banking}

The methodology presented in section~3 is applicable for detecting transactions with the financial crime of fraud. In this paper, we are using this method to detect money laundering in bank daily transactions. To this aim, we are applying both classification and anomaly detection to recognize the transactions which are suspicious for money laundering. The dataset we are using comes from the bank, and it has several features on the transactions. We will use all the features and will not assign any weights to them before applying the models. We want our model to realize the impact of any of these transaction attributes on our research goal, i.e., money laundering prediction.

\subsubsection{Data Source}

The data for our dataset comes from multiple data sources within the bank. Each of these data sources has its feature labels and structure, so the first thing we need to do is extract the standard features and leave the ones specific to one particular type of transaction. For example, "Transaction Branch" might have a different meaning to cash transactions than online banking. In the online banking database, we had a field "Terminal Name" which is the equivalent of "Branch" when dealing with the transactions done in a physical location. There were also some features that would be a duplicate and not necessary for the purpose of this paper. For example, for the transactions done via card, there was a field "Card Number" which would not add any value other than the Account ID for the purpose of this study. Further to that, this feature would be available for just one transaction type and not all of them; therefore, it was decided to delete this feature.
The dataset included data on transactions. Below is a brief description of each feature:

\textbf{Transaction ID:} This is the unique identification number that the banking system will assign to a transaction.

\textbf{Account ID:} This is the unique identification of the account holder, which will help the bank identify which account this transaction may belong to. This is important because a customer who has a history of money laundering is potentially more exposed to doing it again. Customer ID has a risk that the model should not flag people who have an id similar to the id of a suspicious customer. We need to control this with other solutions.

One mitigation plan would be to make customer IDs different from each other to decrease the risk. Another plan would be to flag some customer IDs related to people or companies with high risk and replace customer ID with that feature. This is applicable for both Source customer ID (payer) and Destination customer ID (payee).

\textbf{Account (customer) type:} The customer who is doing the transaction can have different types; it can be an individual or a company doing the transaction. It can also be an association or a trust. This is specifically important because different customer types have different transaction amounts and transaction behavior. What is unusual for an individual might be a recurring character of a company doing its regular transactions.

\textbf{Product Type (Transaction Type):} This field specifies the product type that is engaged in the transaction. There can be several product types in each bank, but the most common ones are cash-in, cash-out, card, direct payment in the bank, cheque-in, cheque-out, new platforms (such as m-banking or internet banking), and global payment.

\textbf{Transaction Code:} Each transaction type can have several methods; for example, card payment can be an online purchase or tapping in a supermarket, or transferring money using an ATM. Each of these makes different codes under each transaction type. For some of the transaction types, only two or three codes are available, whereas, for others like card transactions, there are more than ten different codes.

\textbf{Transaction Branch:} This feature will specify in which branch the transaction took place. This can be a physical location where the transaction has taken place, an ATM or another machine related to a branch, or a code for the payment gateway that the online transaction has taken place.

\textbf{Source Bank:} Source bank specifies the bank to which the source account belongs to.

\textbf{Destination Bank:} Destination bank specifies the bank of the account to which account the transaction has been sent to.

\textbf{Transaction Date/Time:} This specifies the date and time in which the transaction has taken place in. This might also be important to check and see whether the time of a transaction is essential or not. For example, doing a transaction in the middle of the night might be a sign to check whether it is typical or unusual.

\textbf{Transaction Amount:} The transaction amount is the monetary value of the transaction. This is one of the most important features based on the experience and the quotes from subject matter experts. The model will, however, test this to see if this perception is valid or not.

\textbf{Average Amount of transaction in the previous month:} This feature specifies the average amount for all the transactions for that account in the previous month. This is also useful as it will give some insight into the account holder's average transaction amounts.

\textbf{Transaction Currency:} This specifies the currency in which the transaction has taken place, e.g., was it Australian Dollar, or it was Thai Baht, or American Dollar.

\textbf{Credit/Debit Status:} This feature specifies whether the transaction was done as a credit or it was in debit mode.

\textbf{Country of Origin:} This mentions the country which was the source of the transaction and the transaction has originated from. In our dataset, more than ninety-five percent of the transactions are originated from Australia.

\textbf{Country of Destination:} This specifies the country to which the money has been transferred to. This is of particular importance as lots of the financial crimes are related to countries to which the regulations of international baking are not implemented, are on the FATF blacklist, or are subject to terrorism or wildlife trafficking. This is based on the knowledge obtained, and the model will test to see if this assumption is valid or not. In our dataset, more than ninety-five percent of the transactions are sent to Australia.

\textbf{Customer Credit Score:} This feature refers to the credit score of the client who is doing the transactions. Every client has a credit score between 0 and 1, with 0 being the lowest and 1 being the highest. The amount of this is specified by the bank through several measures during years.

\subsubsection{Knowledge Sources}

\begin{figure}[t]
    \centering
    \includegraphics[width=1\textwidth, angle=0]{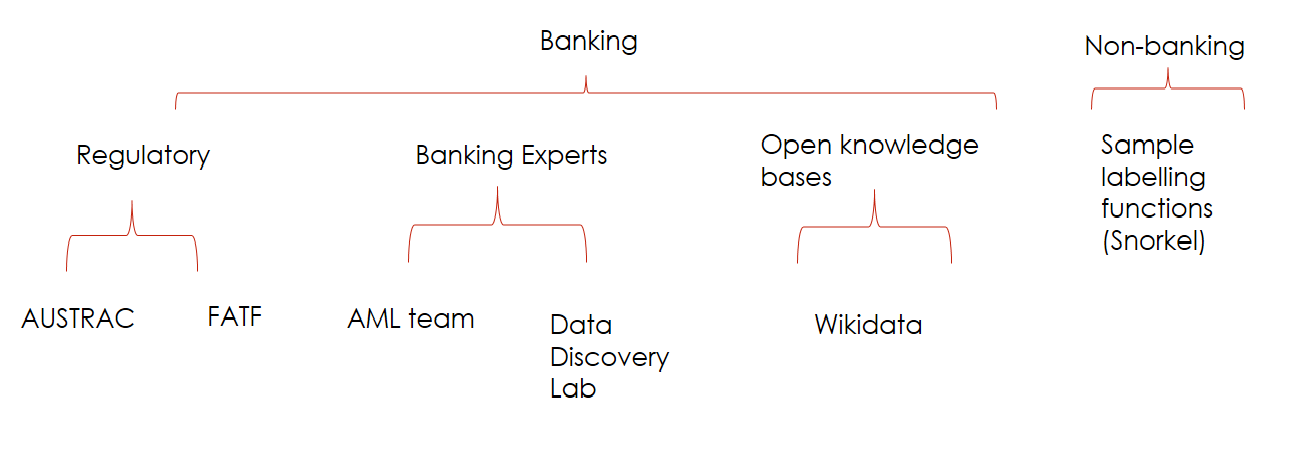}
    \caption{Different Knowledge Sources Used in AML Project.}
    \label{fig: Knowledge Sources.png}
\end{figure}

In this paper, we have different knowledge sources which were taken into consideration. Figure ~\ref{fig: Knowledge Sources.png} shows how multiple sources of knowledge are utilized in this project.

To be specific, we have both banking and non-banking sources as our knowledge sources. For the non-banking area, samples of the labeling functions for the snorkel model coming from academic papers can be a good example. These samples shared a good insight into how we can define our labeling functions and how we can align them to have meaningful results.

In the banking area, we have Regulatory sources, Banking experts, and open source knowledge bases. The Regulatory sources specifically consist of the files and documents from AUSTRAC and FATF as two key sources that define the rules and regulations for money Laundering. These sources have guidelines (both publicly and specific to financial institutions) that provide some rules on what is defined as money Laundering and how to identify it. This is especially important when we want to write our labeling functions, which come from these rules and wordings from regulators. For example, AUSTRAC specifies which countries are mainly considered a thread to commit money laundering for child abuse or which words are related to terrorism activities.

Also, FATF publishes a list of blacklisted countries that support terrorism or have huge drug-organized groups. There are also numerical rules set by these regulatory authorities, which make it a guideline for writing labeling functions. For example, the transaction amount above a certain threshold needs to be investigated. The average amount of transactions in the past $x$ days should not exceed by y percent more than the previous days. These are all defined by the regulatory as a general guide, but financial institutions will refine that based on their internal processes when it comes to practice. In this project, we have used these rules as a source for defining the labeling functions. 

Banking experts were another source of knowledge in this project. As this project is sponsored by one of the Australian largest banks, multiple sessions have been held with several banking experts attending and sharing their knowledge. This knowledge was in two inter-connected areas. The first area was the practical knowledge that has been obtained through years of working in financial crimes teams. The second area was the technical knowledge of the software and systems which are currently being used by the banking industry or can be potentially used to enhance the current rates, such as increasing true positive detection and decreasing false positive ones.

In order to accomplish this aim, two teams were explicitly helpful in this area. One was the anti-money Laundering team with years of experience in financial crimes and fighting money Laundering cases and had their ways of finding their cases and reporting them to AUSTRAC. Their method was mainly rule-based, utilizing the software systems written based on AUSTRAC general guides. The other data gathering source was the data discovery lab which gathered the data from different sources and transformed it into a format that was usable by the project. These sources helped shape the path of the project on the go.

The data itself also comes from different sources. Each transaction type has a data source system, and the format of the data and the fields in each data source are different. A critical step in this project was aggregating all these different data types and shaping them into a format that can be used across the project.

Another source of Knowledge is open knowledge bases such as Wikidata. These are a common source of open data and were explicitly used to refrain from the transaction references and find the words that had a similar meaning to what has been written in the transaction statement. This tool can enrich the data and decrease the possibility of missing a word related to money Laundering solely because a synonym has been used.
For example, we can find out that "kid" can have the same use as "child" to analyze transaction statements to enhance the chance of finding a suspicious transaction.

\subsection{Data Pre-Processing}

After aggregating the databases and before applying machine learning algorithms on the data, we needed to pre-process the data to get proper results. This includes absolute encoding values, standardizing, and augmenting the data.

\subsubsection{Encoding categorical values}

Some of the features that we had were represented in texts; for example, transaction reference, transaction country, source, and destination countries. We needed to convert these values to categorical values in python and have numerical values for these features. Therefore we used ".cat.code" to convert these texts to numerical values so that the model can read the values appropriately.

\subsubsection{Standardising}

The features that were used in the model had different ranges. For example, the transaction amount would vary from 1 dollars to five hundred thousand dollars, whereas the customer credit score was between 0 and 1. Therefore, the data needed to be standardised to prevent the effect of these varieties in range, using the mean and standard deviation of each variable. The python package used for this aim was sklearn.

\subsubsection{Augmenting}

The data used in this paper was skewed as the number of suspicious transactions among all transactions is low and less than 10 percent. This will lead the machine learning models to have improper results. Therefore we applied SMOTE package in python to augment our dataset. This will cause a more balanced dataset which improves the accuracy of ML applications.

\subsection{Auto-labelling the Data}

The initial labeling was done through AML experts in the bank; however, not all the data in our dataset was labeled, and we needed to apply auto-labeling methods to the data. We used the Snorkel package from python to label the rest of the data.

We had 10 percent of our data labeled by banking experts, and by applying the Snorkel model, we labeled the rest 90 percent of the data.

Below are the samples of the labelling functions we used for our Snorkel model (please note that due to confidentiality, the rules have been modified):

\begin{enumerate}
  \item If Transaction-Type = "Cash" And Transaction-Amount > 10000 Then Suspicious = true
  \item If Source-Country-Code is between [Pak,SyR,...,Yem] And Transaction-Amount > 10000 Then Suspicious = true  (This is referring to FATF listed countries)
  \item If Source-Country-Code is between [KEN,...] And Transaction-Amount > 20000 Then Suspicious = true  (This is referring to countries that are subject to wildlife trafficking)
  \item If Destination-Country-Code is between [Pak,SyR,...,Yem] And Transaction-Amount > 10000 Then Suspicious = true
  \item If Destination-Country-Code is between [KEN,...] And Transaction-Amount > 20000 Then Suspicious = true
  \item If Transaction-Reference is between [Special Words List] And Transaction-Amount > 5000 Then Suspicious = true
  \item If Transaction-Category is between [Special Words List - Category] And Transaction-Amount > 5000 Then Suspicious = true
  \item If Source-Customer-Type = "Individual" And Transaction-Amount>1.5*Avg-Amount-of-Transaction-Previous-Month and Transaction-Amount > 10000 Then Suspicious = true
  \item If Source-Customer-Type = "Organisation" Or "Association" Or "Trust" And Transaction-Amount > 2*Avg-Amount-of-Transaction-Previous-Month and Transaction-Amount > 20000 Then Suspicious = true
  \item If Source-Customer-Type = "Individual" And Customer-Credit-Score < 0.05 and Transaction-Amount>20000 Then Suspicious = true
\end{enumerate}

We used that labeled dataset to run our classification model. Moreover, once the anomaly detection application extracts the unusual transactions, we compared the results by our labels to find the accuracy of our anomaly detection model independently. In other words, while anomaly detection is an unsupervised application, we applied the same accuracy indexes, including Precision, Recall, and F1 score, to provide consistency amongst the results of these two applications (classification and anomaly detection). Finally, once we applied the logical AND between the results, we used the same labels produced by banking experts and the Snorkel model to evaluate the final hybrid model. Below is the definition of the mentioned evaluators for the classification model:

    $Accuracy = \frac{TP + TN}{(TP+TN+FP+FN)}$ (Equation~1)




    $Precision = \frac{TP}{(TP+FP)}$ (Equation~2)

    $Recall = \frac{TP}{(TP+FN)}$ (Equation~3)


    $F-measure = \frac{2*Precision*Recall}{Precision+Recall}$ (Equation~4)


In the above formula, TP stands for True Positive, which refers to the model's cases correctly as a suspicious transaction. TN stands for True Negative and refers to the transactions that the model has recognized correctly as not suspicious. FP stands for False Positive and refers to transactions that the model has recognized as money laundering, but they are not a case of money laundering. FN stands for False Negative and refers to the cases that the model has recognized as a typical transaction, but actually could be related money laundering.

\subsection{Classification}

\begin{figure}[tt]
    \centering
    \includegraphics[width=1.0\textwidth, angle=0]{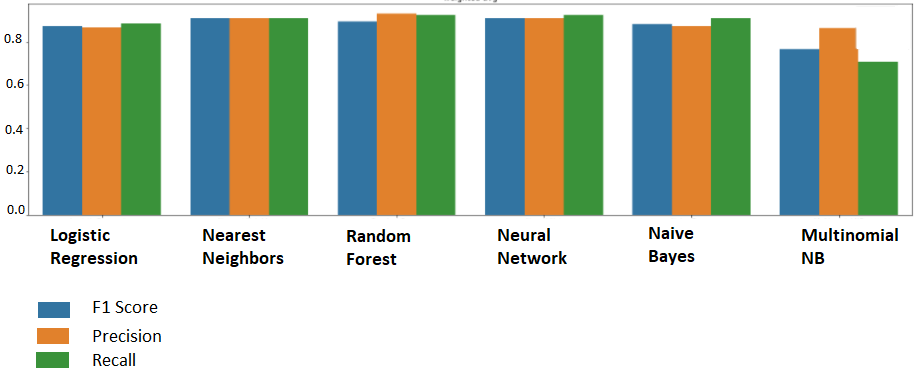}
    \caption{Scores for Classifiers.}
    \label{fig: classification Scores.png}
\end{figure}

After pre-processing our data, we used this data as an input to our binary classification model. We split the data into training and test data with 80 percent for training and 20 percent for the test.
To evaluate which of the classification models have better results on our data, we use Accuracy, Precision, Recall, and F1 Score. These measures are available using sklearn package in python. We applied different classifiers, including Logistic Regression, Nearest Neighbours, Random Forest, Neural Network, Naive Bayes, and Multinomial NB (Multinomial Naive Bayes). 
Table~\ref{fig: Classification Method Results.png} compares the accuracy of applying different classifiers and figure~\ref{fig: classification Scores.png} illustrates this comparison . The results show that a fully connected neural network provides the best level of accuracy. One reason for this can be related to the non-linear boundaries of the dataset. In other words, as multiple factors are influencing the transaction to be suspicious or not, a simple classifier such as logistic regression cannot predict the results with a high level of accuracy, and deep learning provides valid results~\cite{DBLP:journals/cee/NiuXAPBA20}. In terms of implementation time, the neural network took longer to implement, whereas the others were pretty quick. In terms of recall, Random Forest was the best, and Neural Network was the second but close to the first.


\begin{table}[t]
\caption{Classification Method Results.}
    \centering
    \includegraphics[width=0.9 \textwidth, angle=0]{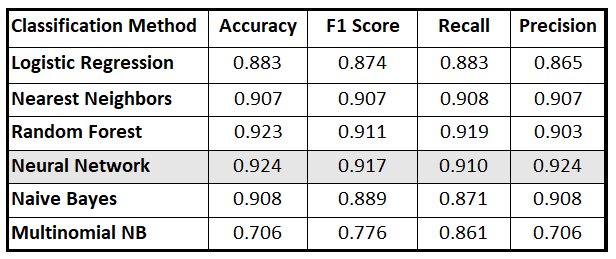}
     \label{fig: Classification Method Results.png}
\end{table}

\subsection{Anomaly Detection}

We applied Anomaly Detection to our data. The idea that we used in evaluating our application of anomaly detection is evaluating this unsupervised machine learning model with some indexes that are usually used to evaluate supervised applications. In other words, once a transaction is identified as an anomaly, i.e., unusual, then we compared the results with our already available labels, which indicate whether or not the transaction is suspicious. Then, by calculating TP, TN, FP, and FN, we measured Accuracy, Precision, Recall, and F1 Score, as shown in table~\ref{table:2}.
Among the algorithms used for anomaly detection, the iForest method had the best result and the highest AUC score, which refers to Area Under Curve~\cite{Vanderlooy2008AAUC}, therefore was selected as the primary anomaly detection method to base the rest of the model onto. A comparison between the results of anomaly detection and classification shows a slightly better performance for classification. However, while the accuracy and precision of anomaly detection are minor compared to classification, anomaly detection improves recall.

\begin{table}[b]
\caption{Anomaly Detection Results.}\label{table comparison}
\centering
 \begin{tabular}{||c c||}
 \hline
  & Result    \\
 \hline
 Accuracy  & 0.893  \\
 \hline
 Precision  & 0.904 \\
 \hline
 Recall  & 0.912 \\
 \hline
 F-1 Score  & 0.907 \\
 \hline
\end{tabular}
\label{table:2}
\end{table}

\subsection{Applying Logical "And" between Two Applications}

In the last step, we applied the logical AND function between the two applications to recognize the transactions that both methods have recognized as suspicious would be considered a suspicious case. The result of this method can be seen in Table~\ref{table:3}. As the results prove, applying logical AND will result in better precision. False-positive cases, i.e., the number of transactions that are identified as unusual but are not suspicious, are reducing. This is aligned with our goal to minimize human intervention by increasing the precision of our model. The results also indicate a slightly better F1 Score after applying logical AND.

\begin{table}[b]
\caption{Applying Logical And between Classification and Anomaly Detection.}\label{table comparison}
\centering
 \begin{tabular}{||c c||}
 \hline
  & Result    \\
 \hline
 Accuracy  & 0.951  \\
 \hline
 Precision  & 0.939 \\
 \hline
 Recall  & 0.899 \\
 \hline
 F-1 Score  & 0.919 \\
 \hline
\end{tabular}
\label{table:3}
\end{table}

\subsection{Discussion}

One of the goals of this research is to minimize the need for human intervention to detect money laundering transactions. As the results show, our method is using two different approaches to extract suspicious transactions. One is classification as a supervised ML model, and the other is anomaly detection as an unsupervised one. Then, by applying logic and reducing the number of transactions that have been extracted as suspicious. In other words, any transaction that is identified as suspicious in this method has much more possibility to be a case of money laundering. Accordingly, the percentage of transactions identified as suspicious and a case of money laundering is rising, which reduces the need for analyst intervention. This is completely aligned with our research goal. The table of results shows a significant improvement in accuracy, F1 score, and precision after applying logical AND, which agrees with our research goal.

\begin{figure}[t]
    \centering
    \includegraphics[width=0.7\textwidth, angle=0]{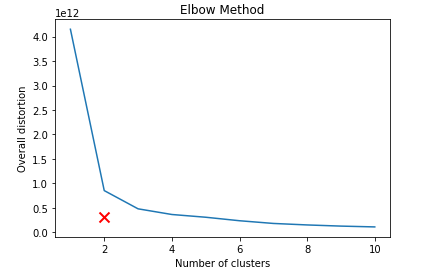}
    \caption{Indication of 2 clusters in Clustering.}
    \label{fig: Clustering Elbow Method.png}
\end{figure}

Finally, to support our findings, we applied a k-means clustering on our dataset to find the segmentation of transactions. As shown in Figure~\ref{fig: Clustering Elbow Method.png}, the optimal number of clusters based on the elbow method is two clusters which are supporting our results in dividing transactions into two groups of suspicious and normal. We aim to combine clustering approaches to provide a visual transaction segmentation in our future works.

\section{Conclusions and Future Work}

\subsection{Conclusions}

Banks and financial institutions are struggling to find and prevent financial crimes.
In this context, Artificial Intelligence is able to  help organizations identify and predict financial crimes.
In this paper, we studied money laundering which is a financial crime that aims to concealing the origins of money.
We discussed intelligent methods to prevent this type of financial crime as the motivating scenario.
We presented a novel approach that aims to extract different features of transactions as raw data from different data islands and enriched them by using knowledge bases (KBs). We applied the process optimization techniques~\cite{DBLP:conf/icsoc/SunBBB15} and fed them into different classifiers to predict suspicious transactions. Our goal was to make improvements in the current prediction rate of money-laundering transactions in terms of precision.

We worked on both supervised and unsupervised machine learning applications for detecting suspicious transactions. We used supervised learning in our classification models to predict suspicious transactions by asking analysts to label the transactions, and develop algorithms that learn from this process and starting to automatically label the data. We also applied the anomaly detection method as unsupervised learning to detect unusual transactions. 
We then combined the two methods, which resulted in more precision resulting in less human work.
We used different features from transactions, such as location and entity type and detected the weight of each in the classification model. We did not pre-weight the features and let our classifier find the impact of each feature on the result. This method applies to any other goal-based transaction classification, such as fraud detection. Obviously, the weight of each feature is changing according to the goal of classification.


We aim to extend our work by using customer features and combining them with transaction features. To achieve this goal, we plan to use data summarization~\cite{DBLP:journals/access/GhodratnamaBZS20}, curation~\cite{DBLP:journals/pvldb/BeheshtiBNT18,DBLP:conf/cikm/BeheshtiBNCXZ17,DBLP:conf/www/BeheshtiTBN17}, and interactive storytelling~\cite{DBLP:conf/www/BeheshtiTB20,DBLP:conf/wise/TabebordbarBB19,DBLP:conf/bpm/BeheshtiSGABYSC18,DBLP:conf/icsoc/AmouzgarBGBYS18} techniques to develop an intelligent feature engineering. 
Another line of related work can focus on intelligent and rule-based~\cite{DBLP:journals/dase/TabebordbarBBB20,DBLP:conf/wise/TabebordbarBBB19,DBLP:conf/icse/TabebordbarB18} recommendations~\cite{DBLP:journals/access/YakhchiBGO020,DBLP:conf/pacis/YakhchiBGO19,DBLP:conf/wise/YakhchiGB18} and trust prediction~\cite{DBLP:journals/access/GhafariBJPMYO20,DBLP:conf/momm/GhafariBJPYJO20,DBLP:conf/momm/GhafariJBPYO19,DBLP:conf/wise/GhafariYBO18a,DBLP:conf/wise/GhafariYBO18} to keep the business analyst aware of potential threats.
We also aim to model the data using graphs modelling~\cite{DBLP:journals/dpd/BeheshtiBM16,DBLP:conf/wise/BeheshtiBNA12,DBLP:journals/cluster/BatarfiSFNBBS15} and querying~\cite{DBLP:conf/bpm/BeheshtiBNS11,DBLP:journals/pvldb/HammoudRNBS15,DBLP:conf/adc/MaamarSBB15,DBLP:journals/dpd/BeheshtiBM16} and apply clustering techniques and combine the result of our classification and anomaly detection with clustering. This is to provide a dynamic dashboard for the banking analyst to have a visible segmentation of each transaction. This will also help the analyst to find similar transactions. Another line of future work could focus on reputation management techniques~\cite{DBLP:journals/compsec/AllahbakhshIBBFB14,DBLP:conf/apweb/AllahbakhshIBBBF13,DBLP:conf/colcom/AllahbakhshIBBBF12} and behavioural analytics~\cite{DBLP:conf/wsdm/BeheshtiHYMG020,DBLP:conf/momm/BeheshtiHY19} to predict the risk-based activities of low profile customers.



\section*{Acknowledgements}
- I Acknowledge the AI-enabled Processes (AIP\footnote{https://aip-research-center.github.io/}) Research Centre and Tata Consultancy Services (TCS) for funding this research.

\bibliographystyle{abbrv}
\bibliography{ms}

\end{document}